\documentclass[letterpaper, 10 pt, conference]{ieeeconf}
\IEEEoverridecommandlockouts
\usepackage[utf8]{inputenc}
\usepackage[T1]{fontenc}    
\usepackage{hyperref}       
\usepackage{url}            
\usepackage{booktabs}       
\usepackage{amsfonts}       
\usepackage{nicefrac}      
\usepackage{microtype}      
\usepackage{xcolor}         
\usepackage{textcomp, gensymb}        
\usepackage{graphicx}       
\usepackage{csquotes}
\usepackage{float}
\usepackage{caption}
\usepackage{subfig}
\usepackage{algorithm}
\usepackage{algorithmic}
\usepackage{amsmath}
\usepackage{amssymb}
\usepackage{multirow}
\usepackage{siunitx}

\title{\LARGE \bf
HDPlanner: Advancing Autonomous Deployments in Unknown Environments through Hierarchical Decision Networks
}

\author{Jingsong Liang$^{1,2}$, Yuhong Cao$^{1}$$^{\dagger}$, Yixiao Ma$^{1,2}$, Hanqi Zhao$^{1}$ and Guillaume Sartoretti$^{1}$
\thanks{$\dagger$ Corresponding author, to whom correspondence should be addressed.}
\thanks{$^{1}$ Authors are with the Department of Mechanical Engineering, College of Design and Engineering, National University of Singapore.
        {\tt\small jsliang@nus.edu.sg, caoyuhong@nus.edu.sg, yixiaoma@u.nus.edu, hanqizhao@u.nus.edu, mpegas@nus.edu.sg}}%
\thanks{$^{2}$ Authors are with the School of Computing, National University of Singapore.
        }
}

\begin{document}
\maketitle
\thispagestyle{empty}
\pagestyle{empty}

\begin{abstract}

In this paper, we introduce HDPlanner, a deep reinforcement learning (DRL) based framework designed to tackle two core and challenging tasks for mobile robots: autonomous exploration and navigation, where the robot must optimize its trajectory adaptively to achieve the task objective through continuous interactions in unknown environments. Specifically, HDPlanner relies on novel hierarchical attention networks to empower the robot to reason about its belief across multiple spatial scales and sequence collaborative decisions, where our networks decompose long-term objectives into short-term informative task assignments and informative path plannings. We further propose a contrastive learning-based joint optimization to enhance the robustness of HDPlanner. We empirically demonstrate that HDPlanner significantly outperforms state-of-the-art conventional and learning-based baselines on an extensive set of simulations, including hundreds of test maps and large-scale, complex Gazebo environments. Notably, HDPlanner achieves real-time planning with travel distances reduced by up to $\SI{35.7}{\%}$ compared to exploration benchmarks and by up to $\SI{16.5}{\%}$ than navigation benchmarks. Furthermore, we validate our approach on hardware, where it generates high-quality, adaptive trajectories in both indoor and outdoor environments, highlighting its real-world applicability without additional training.

\end{abstract}

\section{INTRODUCTION}

Highly stochastic environments, especially partially observable environments, pose a nontrivial challenge for the decision-making of a mobile robot, which requires the capability to reactively adapt the robot's policies through continual interactions with uncertain environments to handle dynamical change within them. This paper investigates two representative and core tasks for the mobile robot in unknown environments: autonomous exploration and navigation. Specifically, autonomous exploration aims to identify the shortest route for entirely covering the unknown environment to classify it into free and occupied areas. In autonomous navigation, the robot is allocated a far destination in an unknown environment and strives to reach it as fast as possible. 

\begin{figure}
  \centering
  \includegraphics[width=0.99\linewidth]{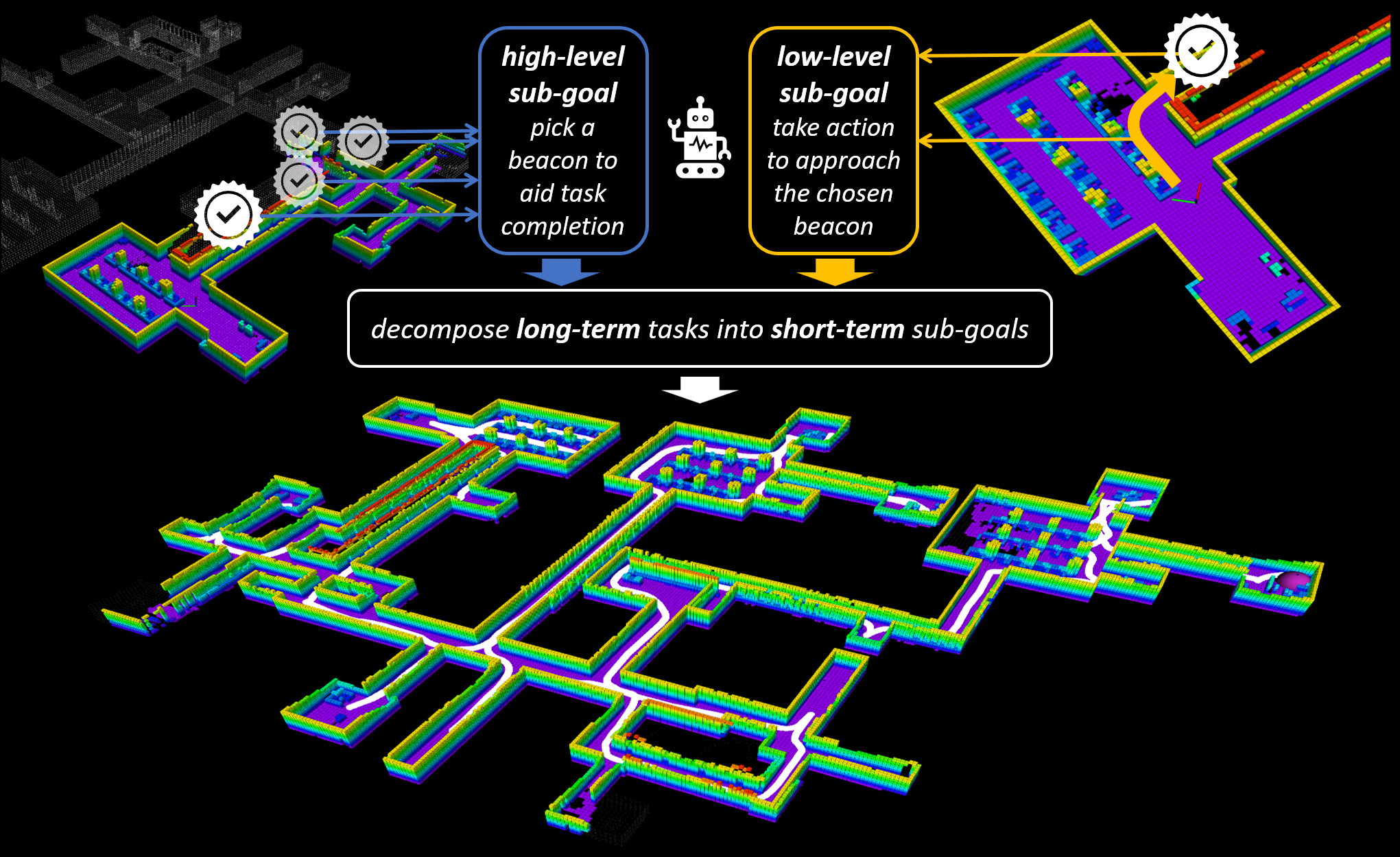}
  \caption{\textbf{Illustration of our proposed hierarchical planning for autonomous deployments.} HDPlanner breaks down long-term objectives for exploration/navigation into short-term sub-goals to efficiently accomplish the tasks: high-level assignments by choosing a beacon (white tick) which indicates informative areas, and low-level reactive planning through consecutive waypoints (yellow arrow) to collect information as approaching the designated beacon.}
  \label{fig: attention vis}
  \vspace{-10pt}
\end{figure}

In these two tasks, the robot maintains a partial \textit{robot belief} about the environment, which is an accumulated observation of the environment. There, the unknown environment will be revealed to the robot along with the robot's movements.
Although the objectives of exploration and navigation are different, they share the same critical underlying mission: to efficiently gain sufficient knowledge of the environment to support task completion.
These tasks involve the non-trivial optimization process for the robot’s trajectory, especially in real-world deployments: To achieve long-term objectives efficiently, the robot must carefully balance exploring the unknown areas with exploiting its existing beliefs for task execution. Meanwhile, online map updates in these two tasks further require real-time re-planning for the robot to guarantee the planned path's effectiveness, making trajectory optimization even more challenging.

While many previous planners tend to optimize the objectives directly, more recent studies found that a hierarchical structure that decomposes the global objective into sub-goals can achieve more efficient overall performance~\cite{haarnoja2018hierarchical}. Notably, the state-of-the-art hierarchical planner for autonomous exploration~\cite{cao2021tare} focuses on the \textit{coarse-to-fine} representation of the robot’s global and local maps, primarily to cope with the computational complexity of representing the environment. However, although~\cite{cao2021tare} prioritizes planning optimal paths based on the current belief, it can not reason about unknown areas to estimate longer-term exploration efficiency. We note such ability is crucial for the final performance of exploration/navigation in unknown environments, as future updates on robot belief may contradict the robot's current plan. 

In this work, we propose HDPlanner, a novel hierarchical planning framework designed for both autonomous exploration and navigation in unknown areas. To the best of our knowledge, HDPlanner is the first hierarchical learning-based approach for these tasks. In particular, we introduce a hierarchical decision structure for both policy and critic networks, empower the robot to learn more efficient policies by \textbf{decomposing a complex long-term task into simpler, shorter-term informative planning}, i.e., beacon assignment and waypoint selection: The \textbf{high-level} planning involves strategic task assignments among potential areas with explicit/implicit information gains to facilitate the overall task completion. The \textbf{low-level} planning entails detailed informative path planning to navigate to the area selected by our high-level policy. 

Compared to state-of-the-art conventional and learning-based methods, our approach shows superior exploration efficiency (\SI{21}{\%} improvement in path length, \SI{14}{\%} in makespan) and navigation efficiency (\SI{16}{\%} improvement in path length, \SI{26}{\%} in makespan), while reducing computing time by \SI{78}{\%} in large-scale, complex Gazebo environments. We also validate HDPlanner and its variants through ablation tests to highlight the benefits of our hierarchical decision network design and proposed training strategies. Finally, we experimentally validate our planner on a ground robot in both indoor and outdoor environments, highlighting its real-life applicability without any additional training.

\section{RELATED WORK}

\subsection{Exploration Planner}
Robotic exploration of unknown environments has attracted the interest of the research community. In frontier-based exploration, a robot is driven to explore the boundaries of an already-explored area (i.e., the frontiers). Therefore, the sequence of frontiers is crucial to avoid visiting repeated frontiers during the exploration. Frontier-based works select the frontier to visit through a gain function \cite{meng2017two,cieslewski2017rapid}, based on cost \cite{osswald2016speeding,julia2012comparison} or utility \cite{niroui2017robot,gonzalez2002navigation} performance metrics. However, such methods only consider short-term efficiency, making them always perform myopic behavior in large-scale environments.

More recent works rely on sampling-based approaches to plan longer-term paths, thus improving the long-term exploration efficiency, such as NBVP \cite{bircher2016receding}, MBP \cite{dharmadhikari2020motion}, and DSVP \cite{zhu2021dsvp}. Notably, Cao et al. proposed TARE \cite{cao2021tare}, a combination of a TSP-based global planner and a sampling-based local planner with a coverage constraint, to optimize the full exploration path over the current belief, which currently exhibits state-of-the-art performance.

With the advancement of machine learning, DRL-based approaches have also been studied to let the robot learn local decision-making that aligns with long-term objectives. Most existing methods applied convolutional neural network (CNN) techniques to process robot exploration maps as inputs \cite{zhu2018deep,li2019deep,xu2022explore}. In particular, our recent work~\cite{cao2023ariadne} introduced attention networks with graph inputs derived from the robot's observation. While this attention-based approach offers notable improvements over other learning-based methods, it, like other DRL methods, struggles with scalability and effectiveness in large, complex environments.
\subsection{Navigation Planner}
Although plenty of methodologies address the problem of robotic navigation with prior maps, limited work has been devoted to the autonomous navigation problem in unknown/partially known environments. Conventional methods for pathfinding in known areas, including search-based approaches (e.g., A* \cite{hart1968formal} D* \cite{likhachev2005anytime}, LPA* \cite{koenig2004lifelong}, and D* lite \cite{Koenig2005D*Lite}) and sampling-based approaches (e.g., Rapidly-exploring Random Tree (RRT) \cite{lavalle2001rapidly}, RRT* \cite{karaman2011sampling}, RRT-connect \cite{kuffner2000rrt}, and Batch Informed Trees (BIT*) \cite{7139620}) can be employed to handle navigation in unknown environment. Generally, they use discrete grids or a graph with node priority to rapidly plan/replan the possible route toward the destination. However, more recent studies \cite{liang2023context,yang2022far} found that these approaches are prone to be inefficient because they lack subtle predictions of the unknown area towards the destination, as well as suffer from high computational burden at replanning.

DRL-based approaches have attracted the attention of the community in recent years. The DRL planners are expected to maximize the long-term return by reactive reasoning about the current belief. Most of the works can be summarized as end-to-end models that focus on local/motion planning. Specifically, the sensory data, e.g., RGBD image \cite{pfeiffer2017perception} and 2D/3D point cloud \cite{tai2017virtual,jin2020mapless}, are fed forward to the model without building the robot’s partial map. However, these end-to-end models are incapable of long-term planning \cite{pfeiffer2017perception,tai2017virtual}, especially in large-scale and complex environments. More recently, our work~\cite{liang2023context} proposed a context-aware global planner for navigation in large-scale unknown environments, which achieved state-of-the-art performance in planning time and success rate metrics. However, these models still struggle to scale to large, complex environments and cannot outperform D* lite in terms of travel distance, suggesting insufficient study to dig out the full potential of DRL.

\section{PROBLEM FORMULATION}

\begin{figure}
  \centering
  \vspace{6pt}
  \includegraphics[width=0.95\linewidth]{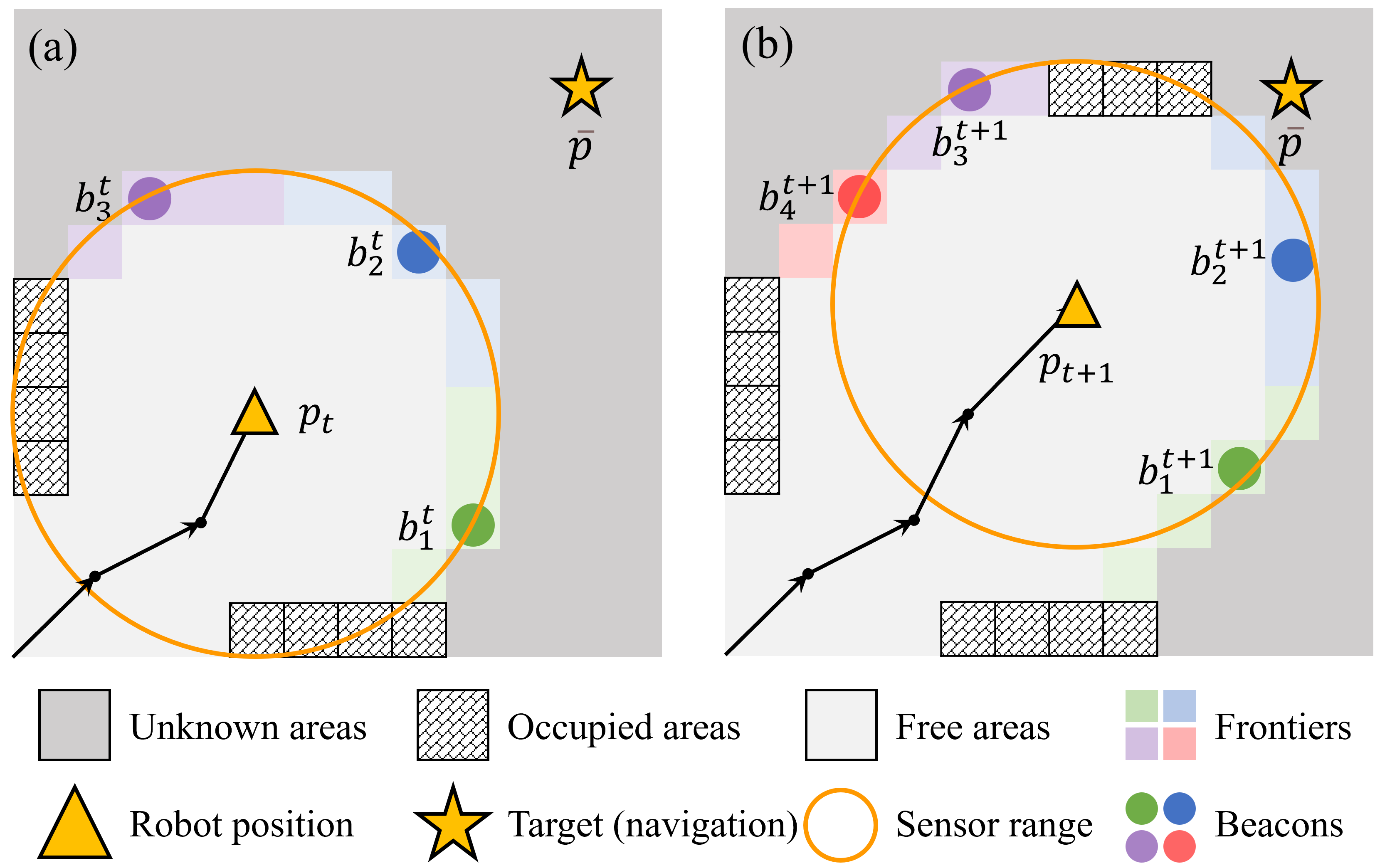}
  \caption{\textbf{Illustration of Observations.} In (a), the robot forms its observation at time step $t$. Specifically, the beacons $b_i^t$ are generated based on the current non-zero utility viewpoints. In (b), upon reaching the new position $p_{t+1}$, the robot's belief expands, leading to an update of the beacons to $b_i^{t+1}$ at time step $t+1$.}
  \label{fig: attention vis}
  \vspace{-10pt}
\end{figure}

\subsection{Autonomous Deployments as RL Problem}
Define $\mathcal{W} \subset \mathbb{R}^2 $ as the overall environment, where it can be divided into free areas $\mathcal{W}_f$ and occupied areas $\mathcal{W}_o$, i.e., $\mathcal{W}_f\cap\mathcal{W}_o=\emptyset$ and  $\mathcal{W}_f\cup  \mathcal{W}_o=\mathcal{W}$. Given a sensor-based observation $o_t$ at each decision-making step $t$, the robot builds and updates a map $\mathcal{M}_t$ which consists of three parts: free areas $\mathcal{M}_t^f$, occupied areas $\mathcal{M}_t^o$, and unknown areas $\mathcal{M}_t^u$. For exploration, the robot should produce a collision-free trajectory $\tau $ which is considered as a sequence of waypoints, to cover the free area  $\mathcal{W}_f$, termed as $\tau =\left \{ p_0, p_1, \dots  \right \}, \forall p_i\in \mathcal{W}_f$. On the other hand, for navigation, the robot is objective to produce a collision-free trajectory $\tau $, which is considered a sequence of waypoints from current position $p_t \in \mathcal{W} $ to the predefined target $\bar{p} \in  \mathcal{W}$. The objective across these two tasks is the same: minimize the cost for the travel distance $dist(\tau)$ or make-span $T(\tau)$ of the given trajectory $\tau$.

We formulate the autonomous deployments as a partially observable Markov decision process (POMDP). A POMDP tuple is denoted as $(\mathcal{S}, \mathcal{A}, \mathcal{T}, \mathcal{R}, \Omega, \mathcal{O}, \gamma)$, where $\mathcal{S}$ represents the state space, $\mathcal{A}$ the action space, $\mathcal{T}$ the state transition function, $\mathcal{R}$ the reward set, $\Omega$ the observation space, $\mathcal{O}$ the observation function, and $\gamma \in \left [ 0,1 \right ]$ the discount factor. During the task, instead of accessing the true state $s^t$, where $\forall{s^t}\in{S}$. the robot bases on its partial observation $o^t$, where $\forall{o^t}\in{\Omega}$, to select and execute an action $a^t\sim\pi(\cdot \mid o^t)$. Then, the next state $s^{t+1}$ is determined by the state transition function $\mathcal{T}(s^{t+1}\mid{s^t},a^t)$. Given finite decision-making step $T$, the robot's objective is to find an optimal policy $\pi^*$ that maximizes the long-term expected return $\mathbb{E} \left [ \sum_{t=1}^{T}\gamma^{t-1}r_t \right ]$, where $r^t\in{R}$.

\subsection{Viewpoint Graph as Observation}
At each decision step $t$, our observation is represented as $o_t=(G_t, S_t)$, where $G_t$ denotes the \textit{viewpoint graph} and $S_t$ the \textit{planning set}. We first construct the viewpoint set $V_t$ in the viewpoint graph, where each viewpoint from this set is uniformly generated based on the robot belief $\mathcal{M}_i$. Then we build an edge set $E_t$. Specifically, each viewpoint builds up to $k$ nearest edges with each other, where the edge set only keeps edges that connect between collision-free nodes, i.e., nodes that are \textit{line of sight} with each other. The construction of viewpoint graph ${G_{t}}=({V_{t}}, {E_{t}})$ keeps the spatial information of current robot belief and eliminates the concerns of collision with obstacles.
Moreover, the planning set $S_t = (\delta_t, u_t, b_t)$ enlarges the observation with three explicit attributes including (1) $\delta_t$: an indicator to record whether a viewpoint has been visited before; (2) $u_t$: viewpoint's utility. The utility at each viewpoint is proportional to the number of observable frontiers from that position, which are defined as the sampled boundaries between explored and unexplored areas. Specifically, observable frontiers refer to those within the viewpoint's line of sight, where lines connecting the viewpoint to frontiers are collision-free and fall within the sensor range; (3) $b_t$: an indicator, named \textit{beacon}.
As formulated in Algorithm \ref{alg: beacon}, beacons inform the robot about optional unexplored areas of interest. They classify all viewpoints with non-zero utility into distinct groups based on their spatial relationships within the current robot belief map. The observation serves as the input of our policy network.

\begin{algorithm}
\caption{Beacon Aggregation}
\label{alg: beacon}
\begin{algorithmic}[1]
\REQUIRE non-zero utility node set $U$, map $\mathcal{M}$, threshold radius $d_{th}$
\STATE Initialize beacon set $V^{b} \leftarrow \emptyset$, covered node set $\overline{U} \leftarrow \emptyset$
\FOR{$ v\in U $}
    \IF{$v \in \overline{U}$}
        \STATE continue
    \ENDIF
    \STATE Find nearby node set $N$ in $d_{th}$
    \FOR{${v}' \in N$ }
        \IF{\text{line}$(v, v^{'})$ is collision free}
            \STATE $\bar{U} \leftarrow {v}'$
        \ENDIF
    \ENDFOR
\ENDFOR
\STATE $V^{b} \leftarrow v$ 
\RETURN beacon set $V^{b}$ based on $U$ and $\mathcal{M}$

\end{algorithmic}
\end{algorithm}

\section{NEURAL NETWORK AND TRAINING}
\begin{figure*}
  \centering
  \vspace{6pt}
  \includegraphics[width=.99\textwidth]{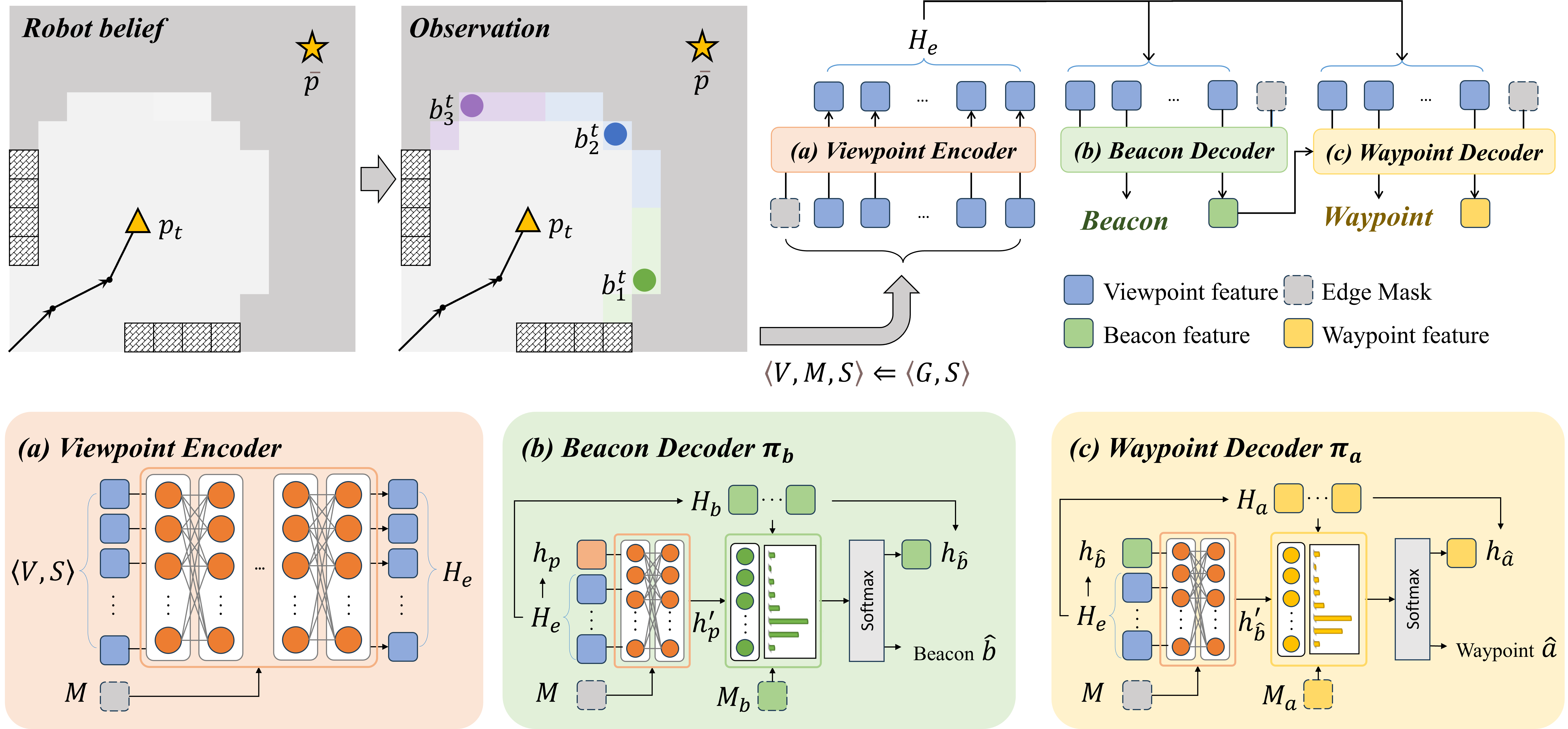}
  \caption{\textbf{Our proposed hierarchical decision networks.} The \textit{viewpoint encoder} first integrates spatial information from the current observation to produce encoded features $\mathcal{H}_e$, where each viewpoint shares its positions along with its planning sets based on the current robot belief. After that, the encoded features of the robot's current position and beacons, $h_p$ and $\mathcal{H}_b$, are used to choose a beacon $\hat{b}$ which indicates the most informative areas, through the \textit{beacon decoder}. The \textit{waypoint decoder} then utilizes the encoded features of the selected beacon and the robot's neighbors, $h_{\hat{b}}$ and $\mathcal{H}_a$, to output a waypoint $\hat{a}$, which indicates a collision-free trajectory started from robot's current position, for informative path planning during the period of approaching the selected beacon.}
  \label{fig: HDPlanner}
  \vspace{-10pt}
\end{figure*}

\subsection{Hierarchical Policy and Critic Networks}
\label{subsec:network}

As demonstrated in Figure \ref{fig: HDPlanner}, our hierarchical decision networks consist of three parts: (1) a viewpoint encoder to learn multiple spatial dependencies among different areas (and predefined target) in the robot belief, (2) a \textit{high-level} beacon decoder to assign robot with the potential most informative areas through the selection of beacon, and (3) a \textit{low-level} waypoint decoder to effectively produce consecutive waypoints to approach the selected beacon while informatively discovering unknown areas.

\subsubsection{\textbf{Viewpoint Encoder}}
Here, we deploy a Transformer-based encoder~\cite{vaswani2017attention} as our viewpoint encoder, which utilizes consecutive attention layers to build the inter-dependencies between every viewpoint and its neighbors at multi-scale perspectives. In each layer, the query $q_i$, key $k_i$, and value $v_i$ are computed as $q_i=f_iW^Q$, $v_i=f_iW^V$, and $ k_i=f_iW^K$, where $W^Q$, $W^V$, and $W^K \in \mathbb{R}^{d\times d}$ are learnable weights. As depicted in \eqref{eq: attention}, the obtained attention $a_{ij}$ is determined with a softmax function, where the similarity $u_{ij}$ is computed via the scaled dot product of the query $q_i$ and key $k_j$. To regulate each viewpoint's access, we introduce an encoder edge mask set $M \in \mathbb{R}^{N\times N}$, which restricts each viewpoint's access to its neighbors. All viewpoint observation vectors $v_{i,t}$ are transformed into $d$-dimensional encoded feature $h_i$ through the viewpoint encoder, where the output encoded feature $h_i$ is calculated as the weighted sum of the unmasked viewpoints values, i.e., $h_i={\textstyle \sum_{j=1}^{N}}a_{ij}v_j$.
\begin{equation} \label{eq: attention}
a_{ij} = 
\begin{cases}
    \frac{e^{u_{ij}}}{\sum_{i=1}^{N} e^{u_{ij}}} & \text{if } M_{ij} = 0\\
    0 & \text{if } M_{ij} = 1
\end{cases}
\text{, where } u_{ij} = \frac{q_i \cdot k_j^T}{\sqrt{d}}
\end{equation}
\subsubsection{\textbf{Beacon Decoder} $\pi_b$}
We extract the encoded feature of robot's position $h_p$ and encoded beacon set $\mathcal{H}_b\in\mathbb{R}^{d\times l_b}$ from the viewpoint encoder, where $l_b$ terms the number of beacons. Here we take $h_p$ as the query source and all the other encoded features as the key-and-value source of a cross-attention layer to output a $d$-dimensional decoded feature of robot's position $h_p^{'}$, which contains the implicit context of the spatial relationships in robot's belief. With decoded feature of robot's position and encoded beacon features fed in as query source and key source respectively, we deploy a pointer layer~\cite{vinyals2015pointer} to directly select the beacon with the largest attention weights as the intermediate sub-goal, i.e., $\hat{b} \sim \pi_b(b_i\in V_b|\mathcal{H}_b, h_p^{'})$.
\subsubsection{\textbf{Waypoint Decoder} $\pi_a$}
The waypoint decoder follows a similar design as the beacon decoder. We first extract the encoded feature of the selected beacon $h_{\hat{b}}\in\mathcal{H}_b$ and the encoded neighbor feature set $\mathcal{H}_a\in\mathbb{R}^{d\times l_a}$, where $l_a$ terms the number of robot's neighbors. Here we take $h_{\hat{b}}$ as the query source and all the other encoded features as the key-and-value source of a cross-attention layer to output a $d$-dimensional decoded feature of selected beacon $h_{\hat{b}}^{'}$. Here, based on the decoded feature of the selected beacon and encoded neighbor feature set, the pointer layer outputs the neighbor with the largest attention weights as the selected waypoint, i.e., $\hat{a} \sim \pi_a(a_i\in V_a|\mathcal{H}_a, h_{\hat{b}}^{'})$.

\subsubsection{\textbf{Training with Hierarchical Critic Network}}
The critic network produces estimations on state-action values, i.e., Q value, from the robot's partial observations, which guides the policy network to learn the optimal policies. Here, we propose a hierarchical critic network, which co-optimizes the estimations on the beacon and waypoint. With this mutual reinforcement of the interrelated objectives, the critic network produces more accurate estimations when interacting with a complex environment and, in return, achieves more adaptive policies. The critic network follows a similar design to the policy network except for these changes: As depicted in Figure~\ref{fig: HDPlanner}, we add a full connection layer to concatenate the decoded feature of selected beacon  $h_{\hat{b}}^{'}$ and the decoded feature of selected waypoint $h_{\hat{a}^{'}}$, into a joint Q value $Q(o, a)$.

During the training process, the critic network outputs the joint Q value to form the critic loss $\mathcal{J(\phi)}$. $\mathcal{J(\phi)}$ as depicted in \eqref{eq: critic loss}, where ${o}'$ is the next observation after taking action $a$ and $V(o)$ is the soft state value. The policy network $\pi_\theta$ is trained to maximize $Q(o, a)$, where the policy loss $\mathcal{J(\theta)}$ is depicted in \eqref{eq: policy loss}. The temperature parameter $\alpha$ is auto-tuned based on the temperature loss depicted in \eqref{eq: alpha loss}, where $\mathcal{H}$ is the predefined target entropy~\cite{haarnoja2018soft, christodoulou2019soft}.

\begin{equation} 
    V(o) =\mathbb{E}_{a\sim \pi_{\theta }(\cdot|o)}\left [Q(o, a)-\alpha\log_{}{\pi_{\theta }(a|o)}\right ]
\end{equation}

\begin{equation} 
    \label{eq: critic loss}
    \mathcal{J}  (\phi )= \mathbb{E}_{o\sim \mathcal{D }} \left [ (Q_{\phi}(o, a)-(r+\gamma (1-d)V({o}' )) )^2\right ] 
\end{equation}

\begin{equation} 
    \label{eq: policy loss}
    \mathcal{J}(\theta ) = \mathbb{E}_{o\sim \mathcal{D }, a\sim \pi_\theta  }\left [  \alpha \log_{}{\pi_\theta(a|o)-Q_\phi(o, a) } \right ]
\end{equation}

\begin{equation}
    \label{eq: alpha loss}
    \mathcal{L}(\alpha) = \mathbb{E}_{a\sim \pi_{\theta} }\left [-\alpha (\log_{}{\pi_{\theta}(a|o)} + \mathcal{H} )\right ]
\end{equation}

\begin{table*}[ht]
    \vspace{10pt}
    \caption{\textbf{Evaluation of exploration planners in simulated maps.} We report the average and variance (shown in parentheses) of the total travel distance, along with the average total exploration steps. The notation "$\downarrow$" implies that a smaller value is preferable, and vice versa (same as below). }
    \label{tab: simulated exploration}
    \centering
    \begin{tabular}{ccccccccc}
        \toprule
                                    & Criteria & TARE & CNN & ARiADNE & Vanilla & Hierarchical & HDPlanner \\ 
        \midrule
        & Distance $(\SI{}{m}) \downarrow$ & $ 246.2 (\pm 58.4)$ & $ 260.0 (\pm 77.3) $ & $239.3 (\pm 66.4)$ & $240.6 (\pm 65.5)$ &  $232.4 (\pm 68.7)$ & $\textbf{221.3} (\pm 61.6)$  \\ 
                                    & Step $\downarrow$ & $ 36.7$ & $ 49.7 $ & $ 30.8 $ & $31.2$ & $ 28.2 $ & $ \textbf{24.5}$  \\
        \bottomrule
    \end{tabular}
\end{table*}
\begin{table*}[ht]
    \caption{\textbf{Evaluation of navigation planners in simulated maps.} We report the average and variance of the total travel distance, the average total navigation steps, as well as the success rates in completing the navigation task.}
    \label{tab: simulated navigation}
    \centering
    \begin{tabular}{ccccccccc}
        \toprule
        & Criteria & D*Lite & BIT & CA & Vanilla & Hierarchical & HDPlanner \\ 
        \midrule
        & Distance $(\SI{}{m}) \downarrow$ & $ 130.2 (\pm 58.7)$ & $ 164.5 (\pm 140.5)$ & $ 140.1 (\pm 98.4)$ & $147.3 (\pm 80.1)$ & $ 135.7 (\pm 76.2)$ & $ \textbf{129.8} (\pm 60.2)$  \\  
        & Step $\downarrow$ & $ 14.1 $ & $ 34.1 $ & $16.1 $ & $22.3$ & $ 14.8 $ & $ \textbf{13.5}$ \\
        & Success Rate $(\SI{}{\%}) \uparrow$ & $ \textbf{100} $ & $88.7 $ & $ 99.3 $ & $97.1$ & $ \textbf{100} $ & $ \textbf{100} $ \\ 
        \bottomrule
    \end{tabular}
\end{table*}
\subsubsection{\textbf{Contrastive Joint Optimization}}
Contrastive learning, shown effectiveness in various computer vision and natural language processing tasks~\cite{robinson2021contrastive}, leverages comparison between similar and dissimilar data pairs to capture representation patterns. Here, we propose a contrastive learning-based optimization for the training process to align the action representation with the expected long-term returns while de-emphasizing those actions that hinder performance, especially in interaction with complex environments. As depicted in Algorithm \ref{alg: contrastive}, we sample observations from the replay buffer $\mathcal{D}$ and build contrastive action triplets $\left \langle \hat{a}, a^+,a^- \right \rangle $ from current action space. Here, $\hat{a}$ denotes the waypoint outputted by the policy network, positive action $a^+$ the action with the highest Q value judged by critic networks, and negative action $a^{-}$ a randomly selected one excluding $\hat{a}$ and $a^+$ in the current action space. To mitigate Q value overestimation, both \textit{Q net} and \textit{target Q net} are engaged in the optimization: With probability $\epsilon$, $a^{+}$ is decided by Q net, otherwise by target Q net. The contrastive loss is finally formulated as \eqref{eq: contrastive loss}, where $f(\cdot)$ denotes the feature mapping of outputted action $f(\hat{a}) \leftarrow h_{\hat{a}}^{'} $, and $m$ denotes a preset margin.
\begin{equation} 
    \label{eq: contrastive loss}
    \mathcal{L}(\theta ) = \left \| f\left ( a^{+} \right ) - f\left ( \hat{a}\right )\right \| -\left \| f\left ( a^{-} \right ) - f\left ( \hat{a} \right )\right \|+m
\end{equation}

\begin{algorithm}
\caption{Contrastive Joint Optimization}
\label{alg: contrastive}
\begin{algorithmic}[1]
\REQUIRE policy network: $\pi_\theta$, Q net: $Q_\phi$, target Q net: $Q_{\bar{\phi}}$, learning rate: $\alpha$

\FOR{epoch $= 1$ to $M$}
    \STATE $\hat{a}\sim\pi_\theta(o)$
    \STATE With probability $\epsilon$, $a^{+} \leftarrow \text{argmax } Q_\phi(o, a)$
    \STATE Otherwise, $a^+ \leftarrow \text{argmax } Q_{\bar{\phi}}(o, a)$
    \REPEAT
        \STATE $a^- \sim \pi_\theta(o)$
    \UNTIL{$a^- \neq a^+$ and $a^- \neq \hat{a}$}
    \STATE Build contrastive triplets $\left \langle\hat{a}, a^+, a^-\right \rangle$
    \STATE $\mathcal{L}(\theta ) \leftarrow f(\hat{a}, a^+, a^-)$
\ENDFOR

\STATE Gradient descent step: $\theta \leftarrow \theta - \alpha \nabla_{\theta} \mathcal{L}(\theta)$
\end{algorithmic}
\end{algorithm}

\begin{figure*}[t]
    \begin{minipage}{.49\textwidth}
        \centering
        \subfloat[Exploration]{\includegraphics[width=\textwidth]{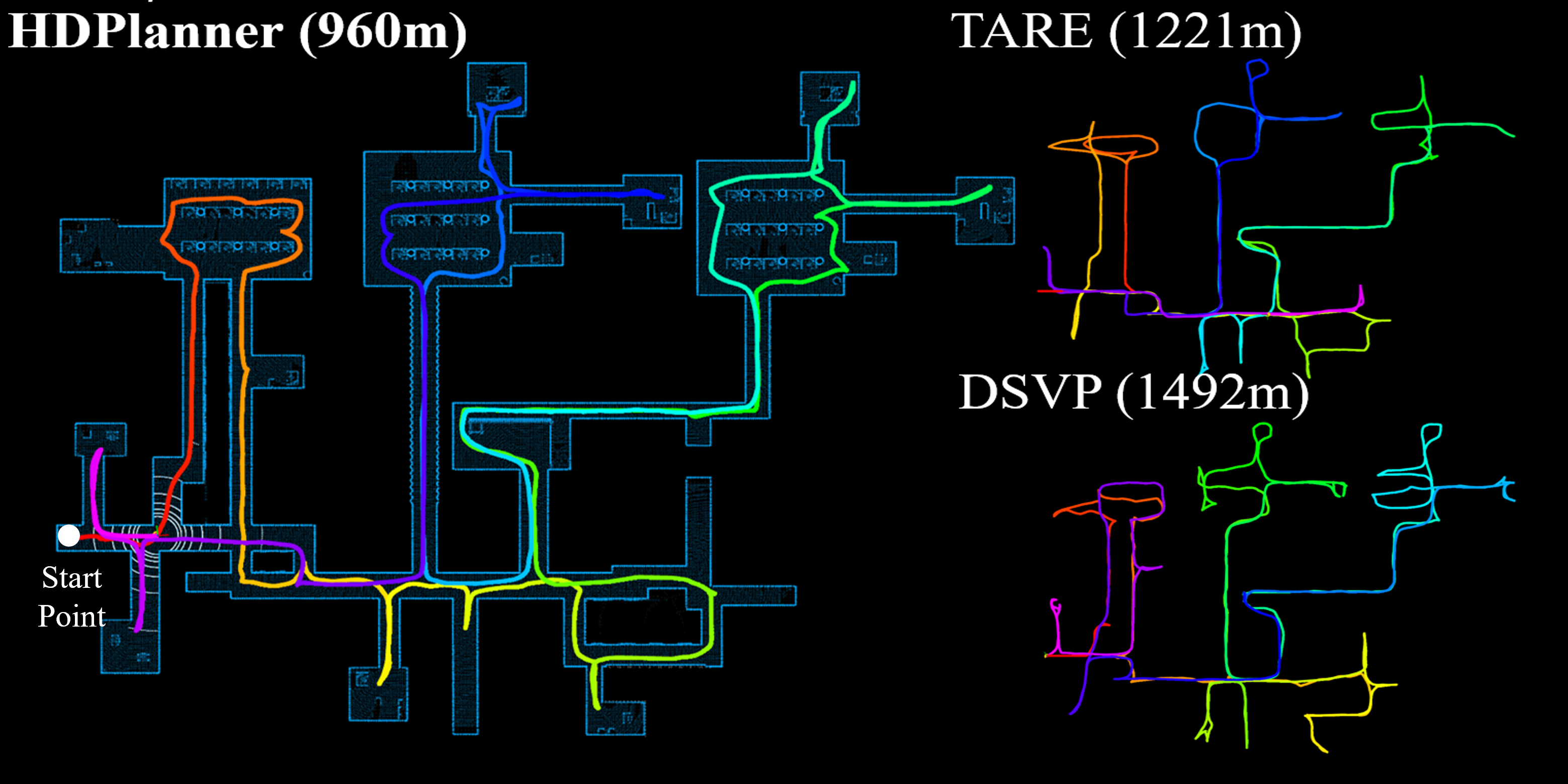}}
    \end{minipage}
    \begin{minipage}{.49\textwidth}
        \centering
        \subfloat[Navigation]{\includegraphics[width=\textwidth]{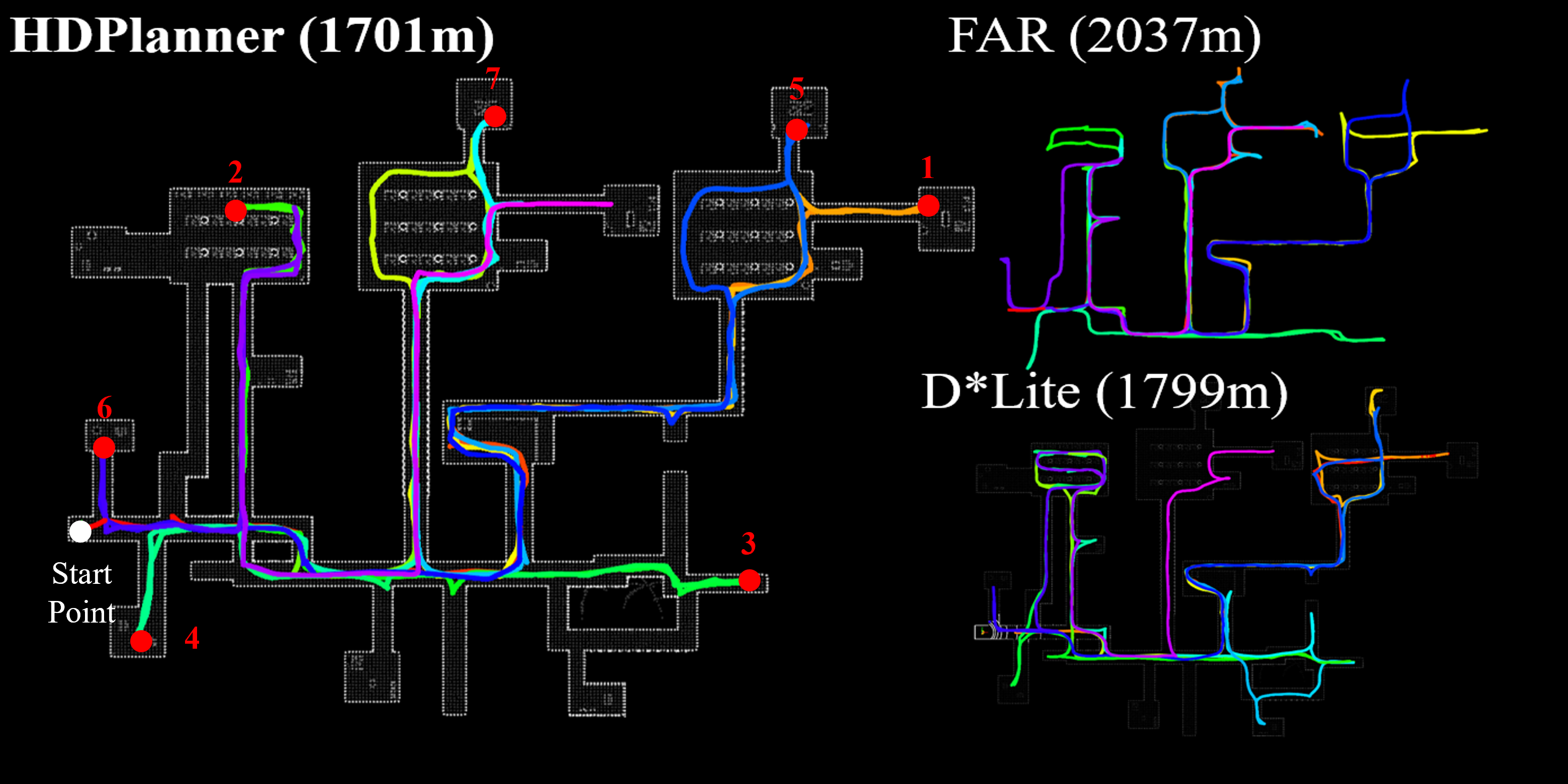}}
    \end{minipage}
    \caption{\textbf{Trajectories comparisons of exploration and navigation planners in large-scale Gazebo simulation.} The colored curve represents the trajectory outputted by the ground vehicle. In (a), each planner starts its exploration from the white dot. In (b), each planner navigates through consecutive predefined targets (red dots). After reaching each target, the planner is reset to ensure the navigation task begins in a fully unknown environment.}
    \label{fig: ros}
\end{figure*}
\subsection{Training Details}

Our model is trained on datasets provided by~\cite{chen2019self} and~\cite{liang2023context}. Here, the robot’s sensor range is set to $\SI{20}{m}$. We uniformly sample $40\times 30$ points for each map. During the robot’s expanding exploration, these sampled points within the free area function as viewpoints. To construct the viewpoint graph, each viewpoint is connected to its 20 closest neighbors, where each connection is collision-free.

HDPlanner is trained on a workstation with an i9-10980XE CPU and two NVIDIA GeForce RTX $3090$ GPUs. We also employ a distributed framework, Ray, to parallelize data collection and training~\cite{moritz2018ray}. The training process takes around 30 hours for coverage. Here, the replay buffer is set to $\SI{10000}{}$ for each episode, the batch size set to $\SI{64}{}$, and the maximum decision-making step limited to $\SI{128}{}$. After each data collection episode, HDPlanner performs four consecutive training iterations. The discount factor $\gamma$ is set to $\SI{0.99}{}$ to balance immediate rewards and long-term returns. We will release our full code upon acceptance.

\section{EXPERIMENTS}
In this section, we conduct experiments across hundreds of simulated maps and large-scale Gazebo environments, incorporating realistic constraints to dig out the superiority of HDPlanner and its component design. Here we introduce two variants of HDPlanner: \textit{Vanilla} and \textit{Hierarchical}. The \textit{Vanilla} variant keeps the viewpoint encoder and waypoint decoder but excludes 1) our hierarchical neural structure in both the policy and critic networks and 2) our contrastive joint optimization during training. The \textit{Hierarchical} variant keeps the hierarchical decision networks while removing the contrastive joint optimization. Additionally, we deploy HDPlanner on a ground vehicle to further validate its effectiveness in real-world scenarios across both indoor and outdoor environments.
\subsection{Validations in Simulated Maps}
We first conduct experiments on an exploration benchmark of 300 maps provided by~\cite{chen2019self} and a navigation benchmark of 500 maps provided by~\cite{liang2023context}. We validate the effectiveness of HDPlanner against exploration baselines including 1) TARE~\cite{cao2021tare}, a conventional hierarchical exploration planner, 2) ARiADNE~\cite{cao2023ariadne}, an attention-based DRL exploration planner, and 3) CNN~\cite{chen2019self}, a CNN-based DRL exploration planner. For comparisons on autonomous navigation tasks, we include 1) D*Lite~\cite{Koenig2005D*Lite}, a search-based navigation planner, 2) BIT~\cite{2015BIT*}, a sampling-based navigation planner, and 3) CA~\cite{liang2023context}, a context-aware DRL navigation planner.

In Table~\ref{tab: simulated exploration}, we report the performance of the exploration approaches. Notably, HDPlanner outperforms all exploration baselines, achieving a $\SI{10.1}{\%}$ improvement over TARE, the state-of-the-art conventional exploration planner, and a $\SI{7.5}{\%}$ improvement over ARiADNE, the state-of-the-art learning-based exploration planner, in terms of travel distance. 
Furthermore, within the family of attention-based exploration planners, Hierarchical significantly outperforms both Vanilla and ARiADNE on complex test sets, underscoring the capability of our hierarchical decision networks to effectively capture dependencies between different areas and decompose complex long-term tasks into simpler, shorter-term sub-goals, i.e., high/low-level informative path planning. In Table~\ref{tab: simulated navigation}, we report the results for navigation approaches. There, HDPlanner consistently outperforms all baseline planners in autonomous navigation tasks. Moreover, Hierarchical shows significant performance gains over Vanilla, further indicating the advantages of our hierarchical decision network design.

We also perform \textit{paired t-tests}, to assess the significance of the comparisons among the different planners. HDPlanner significantly outperforms the state-of-the-art conventional planner, TARE (p-value of \SI{0.00012}{}, well below the standard p-value threshold of 0.05), as well as the state-of-the-art learning-based planners, ARiADNE (p-value of \SI{0.028}{}) and CA (p-value of \SI{0.040}{}), though admittedly with smaller margins for these last two methods. 
In autonomous navigation, we note that, while HDPlanner performs comparably or slightly better than D*Lite (with no significant difference, $p > 0.05$), it achieves significantly lower computational costs ($\sim\SI{0.3}{s}$, including frontier detection, viewpoint graph update, and network inference time) compared to D*Lite ($\sim \SI{1.0}{s}$, both implemented in Python). 

\begin{figure*}[ht]
    \vspace{6pt}
    \begin{minipage}{.15\textwidth}
        \centering
        \subfloat[Ground vehicle]{\includegraphics[width=.95\textwidth]{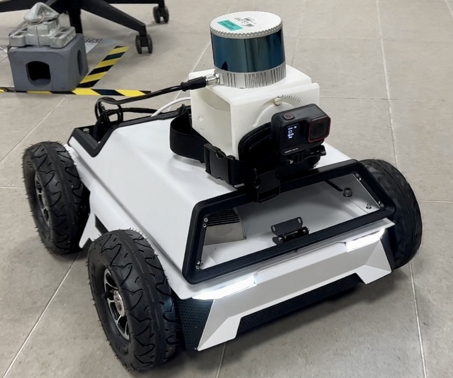}}
        \vspace{-0.2cm}
        \subfloat[Outdoor]{\includegraphics[width=.95\textwidth]{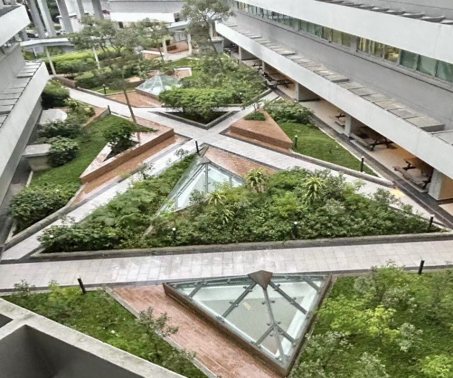}}
    \end{minipage}
    \begin{minipage}{.15\textwidth}
        \centering
        \subfloat[Indoor-1]{\includegraphics[width=.95\textwidth]{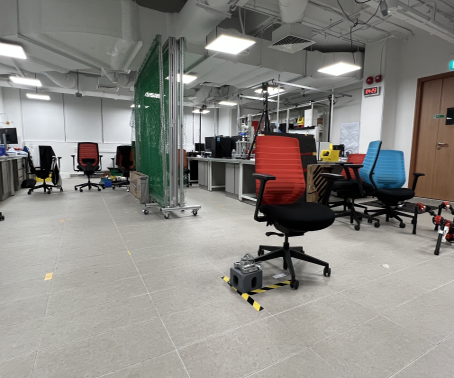}}
        \vspace{-0.2cm}
        \subfloat[Indoor-2]{\includegraphics[width=.95\textwidth]{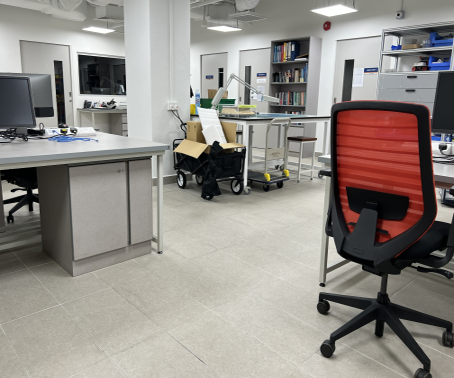}}
    \end{minipage}
    \begin{minipage}{.34\textwidth}
        \centering
        \subfloat[Final point cloud map for exploration task]{\includegraphics[width=\textwidth]{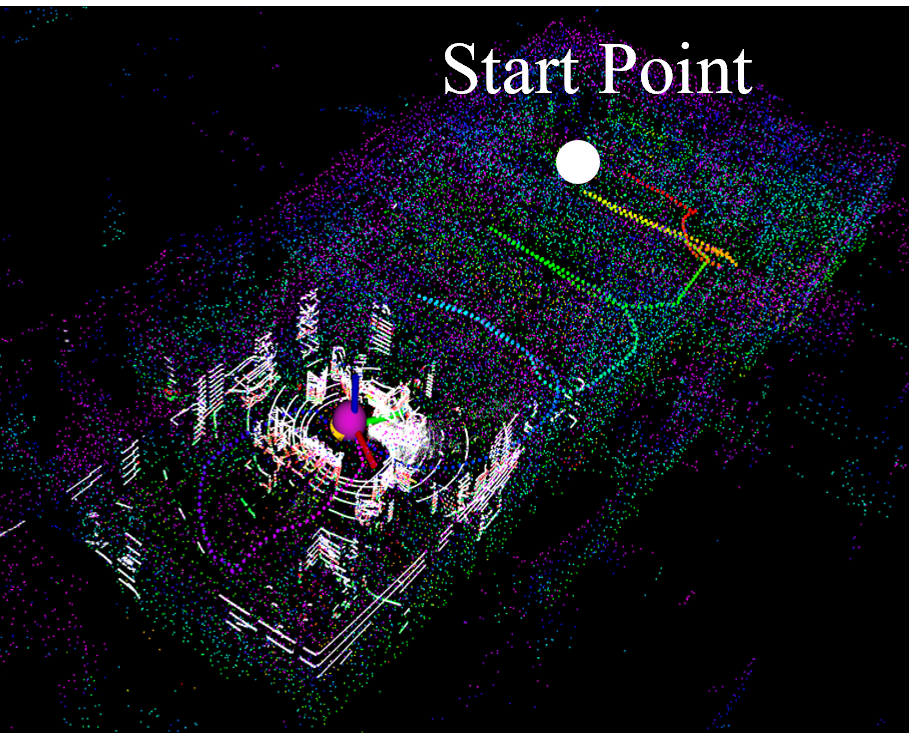}}
    \end{minipage}
    \begin{minipage}{.34\textwidth}
        \centering
        \subfloat[Final Octomap for navigation task]{\includegraphics[width=\textwidth]{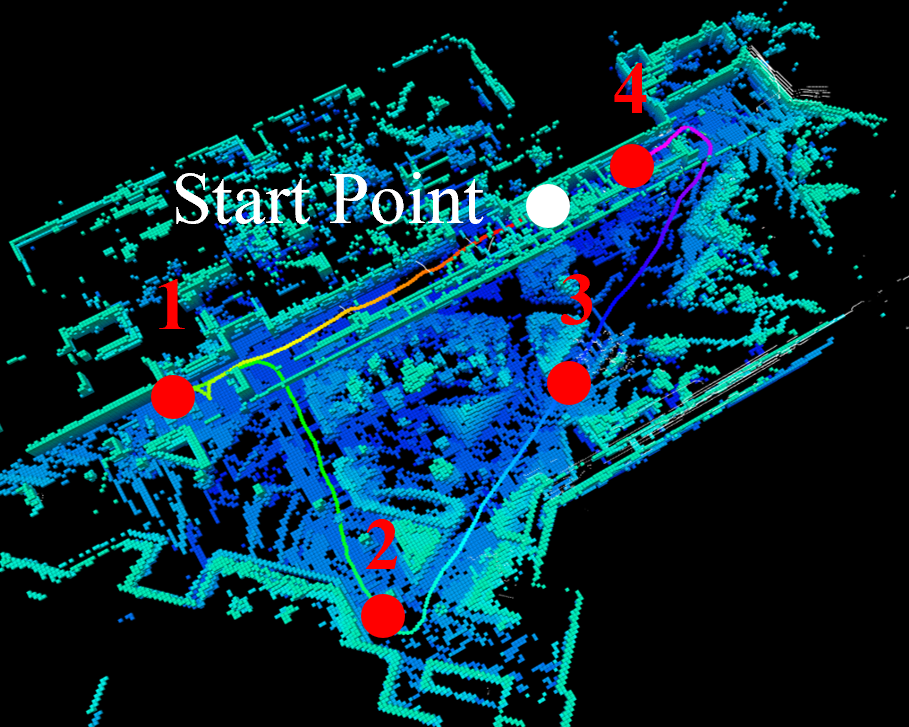}}
    \end{minipage}
    \caption{\textbf{Real-robot experiments.}}
    \label{fig: real robot}
\end{figure*}
\begin{table}[t]
    \caption{\textbf{Comparisons with exploration planners in the large-scale Gazebo simulation}}
    \label{tab: ros exploration}
    \centering
    \begin{tabular}{ccccc}
    \toprule
         &  DSVP & TARE & Hierarchical & HDPlanner\\   
    \midrule
    Distance ($\SI{}{m}$)          $\downarrow$      &  1492&  1221&  1004 &\textbf{960} \\
    Time ($\SI{}{s}$)              $\downarrow$      &  994&   664&  619 &\textbf{571}\\
    Computing ($\SI{}{s}$)         $\downarrow$      &  0.98&  0.25& \textbf{0.21} & \textbf{0.21}\\
    Efficiency ($\SI{}{m^3/m}$)     $\uparrow$        &  3.74&  4.49& 5.33  &\textbf{5.63} \\
    \bottomrule
    \end{tabular}
\end{table}

\begin{table}[t]
    \caption{\textbf{Comparisons with navigation planners in the large-scale Gazebo simulation}}
    \label{tab: ros navigation}
    \centering
    \begin{tabular}{ccccc}
    \toprule
         &  D*Lite & FAR & Hierarchical & HDPlanner\\   
    \midrule
    Distance ($\SI{}{m}$)  $\downarrow$         & 1799 & 2037 & 1874 &\textbf{1701} \\
    Time ($\SI{}{s}$)      $\downarrow$         & 1538 & 1172 & 1158 &\textbf{1134} \\
    Computing ($\SI{}{s}$) $\downarrow$         & 1.10 & 0.41 & 0.25 &\textbf{0.24} \\
    \bottomrule
    \end{tabular}
\end{table}
\subsection{Validations in Large-Scale Gazebo Environments}
In this section, we conduct experiments in a large-scale Gazebo environment ($\SI{130}{m}\times\SI{100}{m}$)~\cite{cao2022autonomous}, designed to challenge the planner's performance under real-world constraints: The planner is deployed on a wheeled differential robot (max speed of \SI{2}{m/s}) equipped with a Velodyne 16-line 3D LiDAR (sensor range of \SI{130}{m}). Here, we introduce extra approaches for more thorough evaluations: 1) DSVP~\cite{zhu2021dsvp}, a dual-stage exploration planner (with a graph-based global planner and a rapid random tree-based local planner), and 2) FAR~\cite{yang2022far}, a visibility graph-based navigation planner. The evaluations are based on metrics from~\cite{yang2022far, cao2021tare} for fair comparisons: 1) \textit{Distance}: total length of the exploration/navigation path, 2) \textit{Time}: total exploration/navigation time, 3) \textit{Computing}: average computing time per step, and 4) \textit{Efficiency}: exploration volume divided by travel distance during exploration task.\\
As shown in Figure~\ref{fig: ros}, HDPlanner consistently outperforms other approaches in autonomous exploration and navigation tasks, which indicates that even in more large-scale, realistic environments, our design of hierarchical decision networks produces more informative path-planning policies. Furthermore, as detailed in Tables \ref{tab: ros exploration} and \ref{tab: ros navigation}, HDPlanner demonstrates significantly more reactive decision-making capabilities, with up to $\SI{78.6}{\%}$ less computing time for exploration and $\SI{78.2}{\%}$ less for navigation (including belief graph update and network inference), even though HDPlanner is implemented in Python and others in C++. While both HDPlanner and Hierarchical outperform other exploration approaches, only HDPlanner surpasses D*Lite in autonomous navigation experiments, as shown in Table~\ref{tab: ros navigation}. Since the hierarchical decision networks are consistent between HDPlanner and Hierarchical, we attribute this performance difference to the introduction of contrastive joint optimization during the training process, which further enhances the state-of-the-art performances of HDPlanner. 

\subsection{Real-World Experiments}
We deploy HDPlanner on a ground vehicle (max speed of \SI{1}{m/s}) equipped with a Leishen 16-line 3D LiDAR (sensor range of \SI{20}{m}). Here, we set the Octomap resolution to \SI{0.2}{m} and the gap between viewpoints to \SI{1}{m}. A local planner~\cite{zhang2020falco} is deployed to output feasible kinematic commands in response to HDPlanner's output waypoints. We evaluate HDPlanner in two scenarios:
\subsubsection{\textbf{Indoor Laboratory}}: In a cluttered indoor environment (\SI{30}{m}$\times$\SI{10}{m}) with obstacles such as chairs and boxes, HDPlanner effectively explores the environment with $6$ minutes, covering a travel distance of \SI{110}{m}. 
\subsubsection{\textbf{Outdoor Garden}}: In an outdoor garden (\SI{40}{m}$\times$\SI{20}{m}) with unstructured terrain and obstacles like trees and shrubs, HDPlanner successfully navigates to consecutive target positions within about $8$ minutes, with a travel distance of \SI{150}{m}. 

As depicted in Figure \ref{fig: real robot}, HDPlanner generates high-quality solutions for both tasks, which demonstrates its adaptability and promising applicability for real-world deployments.

\section{CONCLUSION} 
\label{sec:conclusion}
In this paper, we proposed HDPlanner, a novel DRL-based framework for autonomous deployments in unknown environments. Our hierarchical decision network sequences effective and adaptive decision-making processes to achieve long-term objectives in autonomous exploration and navigation. Furthermore, the integration of our hierarchical critic networks and conservative joint optimization not only brings an efficient training strategy but also enhances the robustness of the planner. We empirically demonstrated that HDPlanner outperforms state-of-the-art benchmarks across various metrics. 
We also validated it in large-scale, convoluted Gazebo environments, as well as through hardware experiments in indoor and outdoor settings, revealing its capability to handle such complex, realistic environments.
Future work will focus on extending HDPlanner to multi-robot autonomous deployments, which will involve establishing cooperative strategies for robot teams to plan efficient paths and effectively interact in highly stochastic environments.

\bibliographystyle{IEEEtran} 
\bibliography{ref}

\end{document}